\documentclass{article}

\usepackage[nonatbib,final]{bdl_2018}

\usepackage[T1]{fontenc}    \usepackage{booktabs}       \usepackage{amsfonts}       \usepackage{nicefrac}       \usepackage{microtype}      
\usepackage[utf8]{inputenc}

\usepackage{cite}

\usepackage[pdftex]{graphicx}
\graphicspath{{images/}}
\DeclareGraphicsExtensions{.pdf,.jpeg,.png}

\usepackage{multirow}
\usepackage{amssymb}
\usepackage[cmex10]{amsmath}

\usepackage{algorithmic}

\usepackage{caption}
\usepackage[font=footnotesize]{subfig}

\usepackage{url}

\usepackage{geometry}

\usepackage[dvipsnames]{xcolor}

\usepackage[
        hidelinks,
        colorlinks=false,
        pdftex,
        pdfpagelabels,
        bookmarks,
        hyperindex,
        hyperfigures,
        linktoc=all,
        bookmarksnumbered=true,
        bookmarksopen=true
    ]{hyperref}
\hypersetup{}

\usepackage{wrapfig}

\usepackage{tikz-cd}

\newcommand{\bs}{\boldsymbol}

\newcommand{\TZK}{\textbf{TzK }}

\newcommand{\Tdomain}{\bs{\mathcal{T}}}

\newcommand{\ki}{{\bs{\kappa}^{i}}}
\newcommand{\T}{{\bs T}}
\newcommand{\fT}{f_{\T}}
\newcommand{\K}{{\bar{{\bs K}}}}
\newcommand{\Ki}{{{\bs K}^{i}}}

\newcommand{\Z}{{\bs Z}}
\newcommand{\Zt}{{\Z}}

\newcommand{\Ci}{{\bs{ \mathcal{C}}^{i}}}
\newcommand{\params}{\theta}
\newcommand{\decparams}{\psi}
\newcommand{\encparams}{\phi}
\newcommand{\M}{\mathcal{M}_{\params}}
\newcommand{\gM}{g_{\M}}
\newcommand{\gMi}{{\gM^{i}}}
\newcommand{\pdec}{p^{dec}_{\decparams}}
\newcommand{\penc}{p^{enc}_{\encparams}}
\newcommand{\pdeci}{p^{dec}_{\decparams^{i}}}
\newcommand{\penci}{p^{enc}_{\encparams^{i}}}

\newcommand{\DKL}[2]{\mathcal{D}_\text{KL}\left(#1\|\, #2\right)}

\newcommand{\E}[2]{\mathbb{E}_{#1}\left[#2\right]}
\newcommand{\DET}[2]{\left|\det \frac{\partial {#1}}{\partial {#2}}\right|}

\newcommand{\comment}[1]{}

\newcommand{\eg}{{\em e.g.}}
\newcommand{\ie}{{\em i.e.}}

\begin{document}

\title{\TZK Flow - Conditional Generative Model}

\author{
Micha Livne (mlivne@cs.toronto.edu)
\and
David Fleet (fleet@cs.toronto.edu)}

\maketitle

\begin{abstract}
We introduce \TZK (pronounced ``task''), a conditional probability flow-based model that exploits attributes (\eg, style, class membership, or other side information) in order to learn tight conditional prior around manifolds of the target observations. The model is trained via approximated ML,
and offers efficient approximation of arbitrary data sample distributions (similar to GAN and flow-based ML), and stable training (similar to VAE and ML), while avoiding variational approximations. \TZK exploits meta-data to facilitate a bottleneck, similar to autoencoders, thereby producing a low-dimensional
representation. Unlike autoencoders, the bottleneck does not limit model expressiveness, similar to flow-based ML. Supervised, unsupervised, and semi-supervised learning are supported by replacing missing observations with samples from learned priors. We demonstrate \TZK by training jointly on MNIST and Omniglot datasets with minimal preprocessing, and weak supervision, with results comparable to state-of-the-art.

\textbf{NOTE:} This workshop paper has been replaced. Please refer to the following work: \url{http://arxiv.org/abs/1902.01893}
\end{abstract}

\section{Introduction} \label{sec:introduction}
\vspace*{-0.15cm}

The quality of samples produced by latent variable generative models continues to improve \cite{Dinh2016a,Kingma2018}. Among the different approaches, variational auto-encoders (VAE) use a variational approximation to the posterior distribution over latent states given observations \cite{Kingma2013,Rezende2015,Kingma2016}. One can also formulate auto-encoders with an adversarial loss, effectively replacing KL divergence in VAE with a discriminator \cite{NIPS2014_5423, Makhzani2015, Makhzani2018}. More recently, flow-based models have produced striking results \cite{Kingma2018}.  They learn invertible mappings between observations and latent states using maximum likelihood (ML) \cite{Germain2015,Kingma2018,Dinh2014,Dinh2016a}, but invertibility entails computation of the determinant of the Jacobian, which is $\mathcal{O}\left(D^3\right)$ in the general case for $D$-dimensional data. By restricting model architecture so the Jacobian is triangular/diagonal, the computational cost becomes manageable (e.g., by partitioning input dimensions \cite{Kingma2013,Germain2015} or with autoregressive models \cite{Papamakarios2017,Oliva2018}). Continuous normalizing flows \cite{Chen2018,Grathwohl2018} trade such memory-intensive flow-based models, with bounded run-time, for calculation-intensive models with bounded memory requirements (solving invertible ODE problem).

Probability flow-based models allows direct parameter estimation via maximum likelihood (ML). Inspired by such models, we introduce \TZK (pronounced ``task''), a conditional probability flow-based model that exploits attributes (\eg, style, class membership, or other side information) in order to learn tight conditional priors around manifolds of the target observations. 
A \TZK model assumes multiple independent attributes (``knowledge''), so new attributes can be added on-line to an existing model. Compared with models trained with variational inference (VI), \TZK scales well to high-dimensional data, without limiting the expressiveness of the model (via a bottleneck), much like flow-based ML parameter estimation. \TZK also presents a unified model for supervised and unsupervised training, by replacing missing observations with samples from current learned priors. Compared with GAN-based training \cite{NIPS2014_5423,Grover2017a}, \TZK offers stable training, which benefits from optimal (converged) discriminators, without the ad hoc use of regularization in GAN \cite{Arjovsky2017}. We demonstrate \TZK by training jointly on multiple datasets with minimal preprocessing, and weak supervision, while maintaining the ability to sample from each dataset independently.

\section{\texorpdfstring{\TZK}{TzK} Formulation} \label{sec:formulation}

\begin{figure}[t]
    \centering
    \begin{minipage}[b]{0.4\textwidth}
    \centering
    \includegraphics[width=1.0\textwidth]{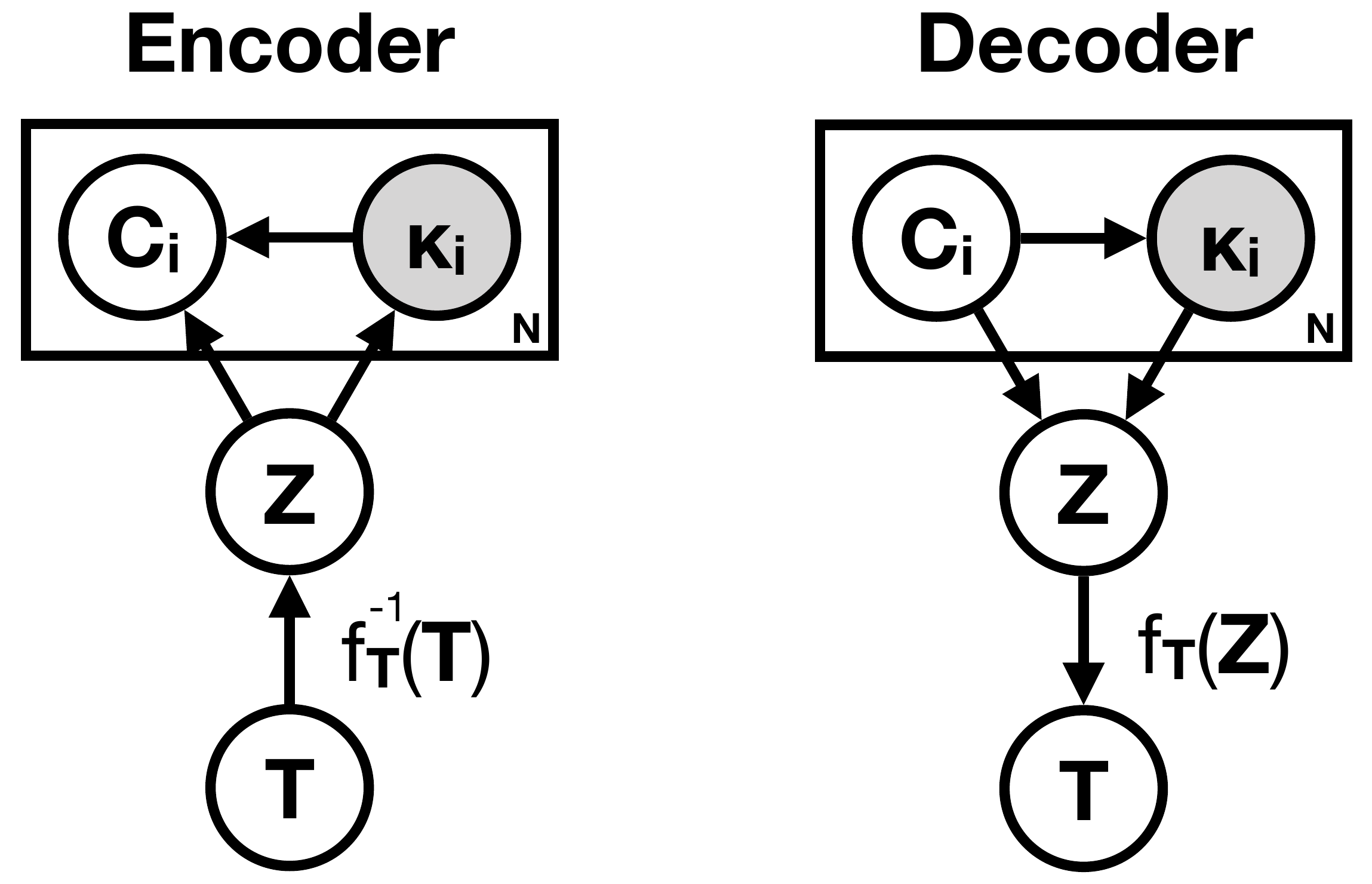}
    \\
    \vspace*{0.1cm}
    \textbf{(a)}  \TZK graphical model
    \end{minipage}
    \begin{minipage}[b]{0.4\textwidth}
    \centering
    \includegraphics[width=1.0\textwidth]{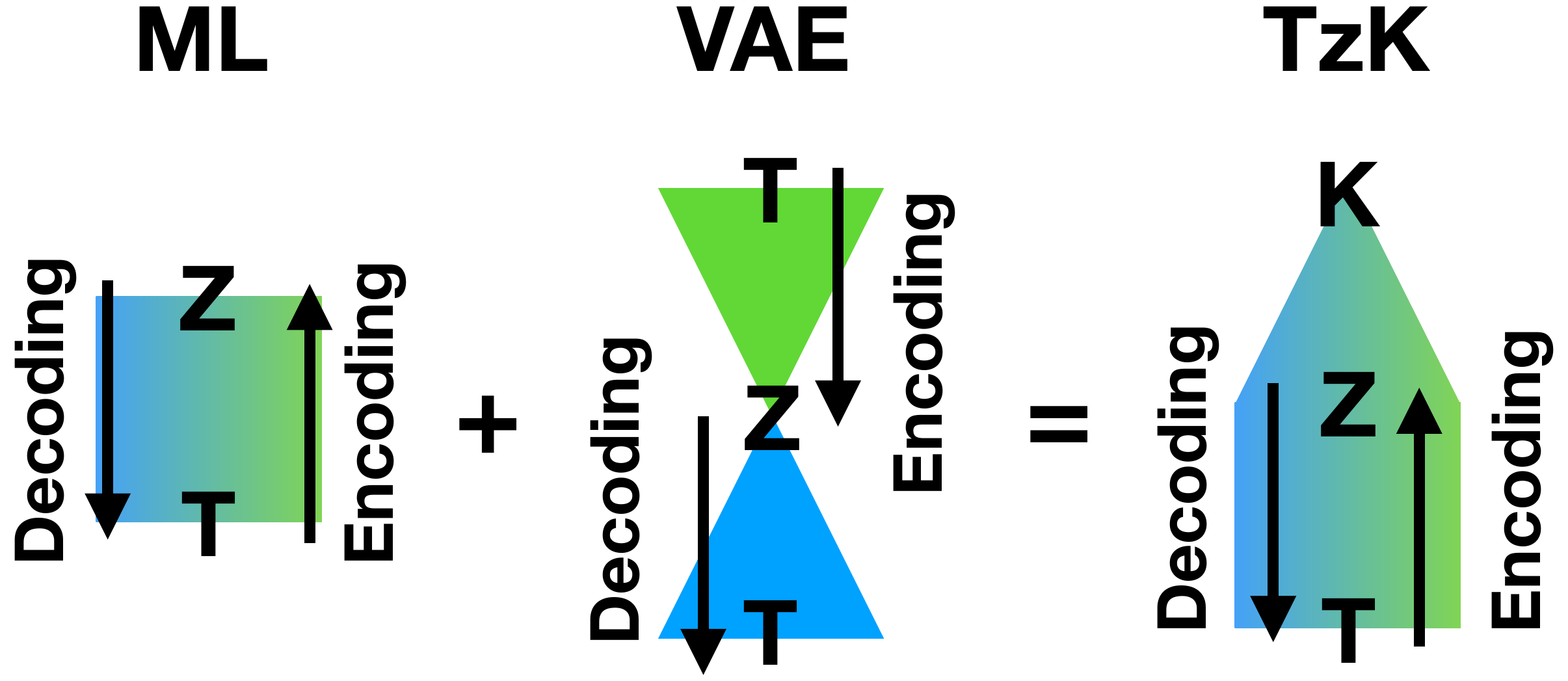}
    \\
    \vspace*{0.1cm}
    \textbf{(b)} \TZK parameters estimation
    \end{minipage}

    \vspace*{0.2cm}

    \caption{\textbf{(a)} \TZK is formulated via a cycle consistent dual encoder/decoder representation of an abstract joint distribution over observations and knowledge types (\eg class label). \textbf{(b)} Comparison with VAE and flow-based ML parameter estimation. Colors represents \textbf{\textcolor{LimeGreen}{encoding}}, and \textbf{\textcolor{Cerulean}{decoding}}. \TZK parameter estimation differs from VAE by placing the bottleneck as a conditional prior $p\left(\T|\Ki\right)$, without limiting the model expressiveness w.r.t.\ $\T$. As a result, \TZK can be optimized directly (similar to ML), and does not require the optimization of a lower bound over an approximated encoder. \TZK differs from ML by incorporating bottlenecks (\ie, conditional priors) to capture manifolds in $\Tdomain$, and thus can be used for semantic dimensionality reduction, like VAE. The use of flow $\T=\fT\left(\Zt\right)$ allows parameter sharing between the various $p\left( \T | \Ki \right)$ and captures the similarity between the different knowledge types.}
    \label{fig:tzk-model}
\end{figure}

A \TZK model comprises an observation $\T \in \Tdomain$, a random variable with probability flow of a corresponding latent variable $\Zt \in \mathbb{R}^D$ \cite{Rezende2015}. Here,
$\Zt$ is mapped to $\T$ through a smooth invertible mapping $\fT~:~\mathbb{R}^D \rightarrow \mathbb{R}^D$ s.t. $\T=\fT\left(\Zt\right)\label{eq:T-flow}$. $\fT$ transforms a given distribution $p\left(\Zt\right)$ (\eg, Logistic or Normal) to the task distribution $p(\T)=p(\Zt)\DET{\fT}{\Zt}^{-1}$.
When $\fT$ is not expressive enough, it is unlikely that the probability density of observed $\fT^{-1}\left(\T\right)$ will match $p\left(\Zt\right)$. As a result, sufficiently expressive models tend to be memory-intensive, with slow training. However, in many cases we would like to sample from a specific manifold of $\T$ (\eg, associated with a style attribute or class label).
We therefore learn distributions over $\T$ conditioned on \textit{knowledge} (or side-information), which also acts as a bottleneck.
We define \textit{knowledge} of type $i$ as a joint discrete/continuous random variable $\Ki= ( \ki \in \{0,1 \} ,\Ci \in \mathbb{R}^{C^i} )$. $\ki$ is a latent binary variable, where $\ki=1$ indicates the presence of knowledge $i$ for a corresponding observation $\T$, and $p\left(\ki=1|\cdot\right) \in \left[0,\ldots,1\right]$ is a discriminator. $\Ci$ denotes knowledge characteristics, and is observed. $\Ki$ specifies a semantic prior over the observation space $\Tdomain$. We define all knowledge types to be independent (Fig. \ref{fig:tzk-model}a). This design choice enables support in unknown number and type of knowledge, allows knowledge overlap (\ie, multiple $\Ki$ can exist in the same sample $\T$, as opposed to supervised 1-hot class label vector), and facilitates on-line acquisition of new knowledge types (\eg, classes).

\paragraph{Example}: $\T \in \mathbb{R}^{1024}$ is a $32\times32$ image, the knowledge $i=\text{``A''}$ denotes the presence of letter ``A'' in an image ($\ki = 1$), and $\Ci \in \mathbb{R}$ represents the associated style. 

\paragraph{Motivation:} A \TZK model facilitates a joint representation of conditional generators, without the requirement for a-priori knowledge about the nature of the condition which is typically required (\eg, number of classes in a conditional VAE). It supports training of multiple labels per observation, while still maintaining the ability to sample from each label independently. The model allows one to incrementally add new side-information (knowledge) in an on-line fashion without having to retrain from scratch with a fixed number of label categories for example.

\newpage
\paragraph{Formulation}: A \TZK model is defined via a cycle consistent dual encoder/decoder representation of an abstract joint distribution over observations $\T$ and knowledge type $\Ki$ (see Fig.\ \ref{fig:tzk-model}a).

\begin{eqnarray}
\penci\left(\T,\Ki\right) &=& \penc\left(\Ki|\T\right)\, \penc\left(\T\right)
~=~\penc\left(\Ci|\T,\ki\right)\penc\left(\ki|\T\right)\penc\left(\T\right)
\\
\pdeci\left(\T,\Ki\right) &=& \pdec\left(\T|\Ki\right)\, \pdec\left(\Ki\right)
~=~\pdec\left(\T|\Ci,\ki\right)\pdec\left(\ki|\Ci\right)\pdec\left(\Ci\right)
\end{eqnarray}

where $\penc\left(\ki=1|\T\right),\pdec\left(\ki=1|\Ci\right)$ are discriminators of binary variable $\ki$, and $\Zt$ is marginalized out exploiting the probability flow deterministic mapping. 

If one were to learn encoders and decoders for each knowledge type independently, then one fails to capture similarities between different kinds of knowledge. For instance, in MNIST dataset the handwritten digits ``1'' and ``7'' can have rather similar images. However, capturing such similarity is hard, and typically requires to know in advance the number and relationships of the different kinds of knowledge. In other words, combining multiple independent generators requires joint training, and thus is typically embedded into the model (\eg, the number of classes has to be known a-prior in a multi-class classification model).

We would like to allow on-line acquisition of unknown number of knowledge types. In addition, we would like an explicit formulation for a decoder (generator) which is conditioned on an arbitrary subset of knowledge types (\ie, compositionality between knowledge types). To achieve that, we propose the following formulation for a \TZK objective. First, we define the joint encoder and decoder likelihoods over $\T$, and multiple knowledge
types, $\K = \{ \Ki \}_{i=1}^N$, as follows:
\begin{eqnarray}
\penc\left(\T,\K\right) &=& p\left(\K|\T\right)\, p\left(\T\right)
\\
\pdec\left(\T,\K\right) &=& p\left(\T|\K\right)\, p\left(\K\right)
\end{eqnarray}
where $\phi$ and $\psi$ are model parameters. Finally, $p\left(\K\right),p\left(\K|\T\right)$ are further factorized with Bayes' rule into individual mappings for different knowledge types,
\begin{eqnarray}
\penc\left(\T,\K\right) &=& \penc\left(\T\right)\prod_{i}\left(\penc\left(\Ci|\T,\ki\right)\penc\left(\ki|\T\right)\right)
\\
\pdec\left(\T,\K\right) &=& \frac{1}{\penc\left(\T\right)^{N-1}}\prod_{i}\left(\pdec\left(\T|\Ci,\ki\right)\pdec\left(\ki|\Ci\right)\pdec\left(\Ci\right)\right)
\end{eqnarray}

with the assumption that knowledge types are independent given $\T$, and have independent priors.

Next, we would like to capture similarity between all existing knowledge types. We do that by
combining $\penc$ and $\pdec$ into a single joint likelihood with $\params = \left\{\encparams,\decparams\right\}$:

\begin{equation}
\M\left(\K,\T\right) ~=~
\left[ \,\penc\left(\T,\K\right) \, \pdec\left(\T,\K\right)\, \right]^{\frac{1}{2}}
\label{eq:knowledge-consistency}
\end{equation}

Of course, for $\M$ to be a valid distribution,  $\penc$ and $\pdec$ must 
represent the same joint distribution. This condition is enforced with the inclusion of a constraint that the KL divergence $\DKL{\pdec\left(\T,\Ki\right)}{\penc\left(\T,\Ki\right)}$ for all knowledge type $i$ should be 0. When the constraint holds for all $i$, the joint encoder/decoder distributions are the same, and $p\left(\T\right)$ accurately accounts for similarity between multiple kinds of knowledge, since 
\begin{equation}
    \penc\left(\T\right) ~=~ \E{\hat{\K}\sim\pdec\left(\hat{\K}\right)}{\pdec\left(\T|\hat{\K}\right)}
\end{equation}
for any subset $\hat{\K}\in \K$ of existing knowledge types $\K$. We refer to $\M$ as the  \textit{knowledge consistency}  of the model to emphasize the joint capability of the model to ``understand'' (encode) and ``express'' (decode). We refer to the constraint the \textit{knowledge gap}, 
\begin{equation}
    \gMi\left(\params\right)\,\equiv\,\DKL{\pdec\left(\T,\Ki\right)}{\penc\left(\T,\Ki\right)}\,=\,\E{\T,\Ki\sim\pdec\left(\T,\Ki\right)}{\log\frac{\pdec\left(\T,\Ki\right)}{\penc\left(\T,\Ki\right)}}
    \label{eq:knowledge-gap}
\end{equation}
as it represents validity of the assumption that the knowledge consistency $\M\left(\K,\T\right)$ is a distribution.When the knowledge gap $\gMi$ is 0, the model allows accurate (as opposed to approximated) compositionality of all possible subsets of acquired knowledge types. A larger knowledge gap represents approximated compositionality, where interpolating (marginalizing) over multiple knowledge types is inaccurate. Investigating compositionality is out of the scope of this paper. The flow $\T = \fT\left( \Z \right)$ resembles the approximated encoder in VAE, and in fact can be any valid flow, including identity. In such a case, \TZK operates similarly to a conditional VAE (CVAE).  When we have no knowledge, the knowledge consistency $\M$ is exactly $p (\T )$.

\paragraph{Objective}: Taking expectation over $\log \M\left(\K, \T\right)$ 
w.r.t. $\T,\K$, we obtain the \textit{observed knowledge consistency}:

\begin{equation}
\E{\K,\T}{\log (\M (\K,\T ) )} \,=\,
-H\left(\T\right)+
\frac{1}{2}\sum_{i}
I\left(\Zt;\Ki\right) - H ( \Ki ) -  \E{\T}{H ( \Ki|\T )}
\label{eq:observed-log-knowledge-consistency}
\end{equation}

where $H(\cdot)$ denotes entropy, and $I\left(\cdot;\cdot\right)$ is mutual information.

\paragraph{Optimization:} We optimize $\params$ by maximizing Eq. \eqref{eq:observed-log-knowledge-consistency} subject to the 
constraint $\penc = \pdec$:
\begin{equation}
\max_{\theta}
\E{\K,\T}{\log\left(\M\left(\K,\T\right)\right)}
-\sum_{i}\lambda_{i} \cdot \gM i\left(\params\right)
\end{equation}
where $\lambda_{i}$ are Lagrange multipliers, $\params$ are the parameters (weights) in $\M$, and $\gMi$ is defined in Eq. \eqref{eq:knowledge-gap}. Even when the knowledge gap constraint in Eq. \eqref{eq:knowledge-gap} does not hold, its presence encourages the model to capture inter-knowledge similarity. In that case, the value of the knowledge gap can be used as an indicator of the violation of the assumption $\penc = \pdec$.

\textbf{Training:} In practice, we approximate expectations w.r.t. $\T,\{\Ci \}_{i=1}^N$ with one sample, which combined with the reparametrization trick and Monte Carlo (MC) approximation yields an unbiased low variance gradient estimator \cite{Rezende2014}. Note that we take exact expectation w.r.t.\  binary variables $\{\ki \}_{i=1}^N$. \TZK support a unified model for supervised, semi-supervised, and unsupervised training. We use supervised samples for MC expectation approximation over continuous variables $\T, \{\Ci \}_{i=1}^N$. When an observation is missing (\eg, unsupervised $\Ci$), we sample from the corresponding conditional prior (\eg, $p(\Ci|\T,\ki )$). Semi-unsupervised training occurs when $\Ci$ is never given externally, while $p\left(\ki|\T\right)$ is. Similarly, this procedure works for $\T$, where we sample from the corresponding $p( \T | \Ci, \ki )$. In such cases we ``freeze'' the discriminators $p (\ki=1|\cdot )$, similar to GAN training, effectively exploiting knowledge already encoded in the discriminators. Full derivation of the procedure requires more space than is available here.

\section{Experiments} \label{sec:experiments}

\begin{figure}[t]
    \centering
    \begin{minipage}[b]{0.49\textwidth}
    \centering
    \includegraphics[width=1.0\textwidth]{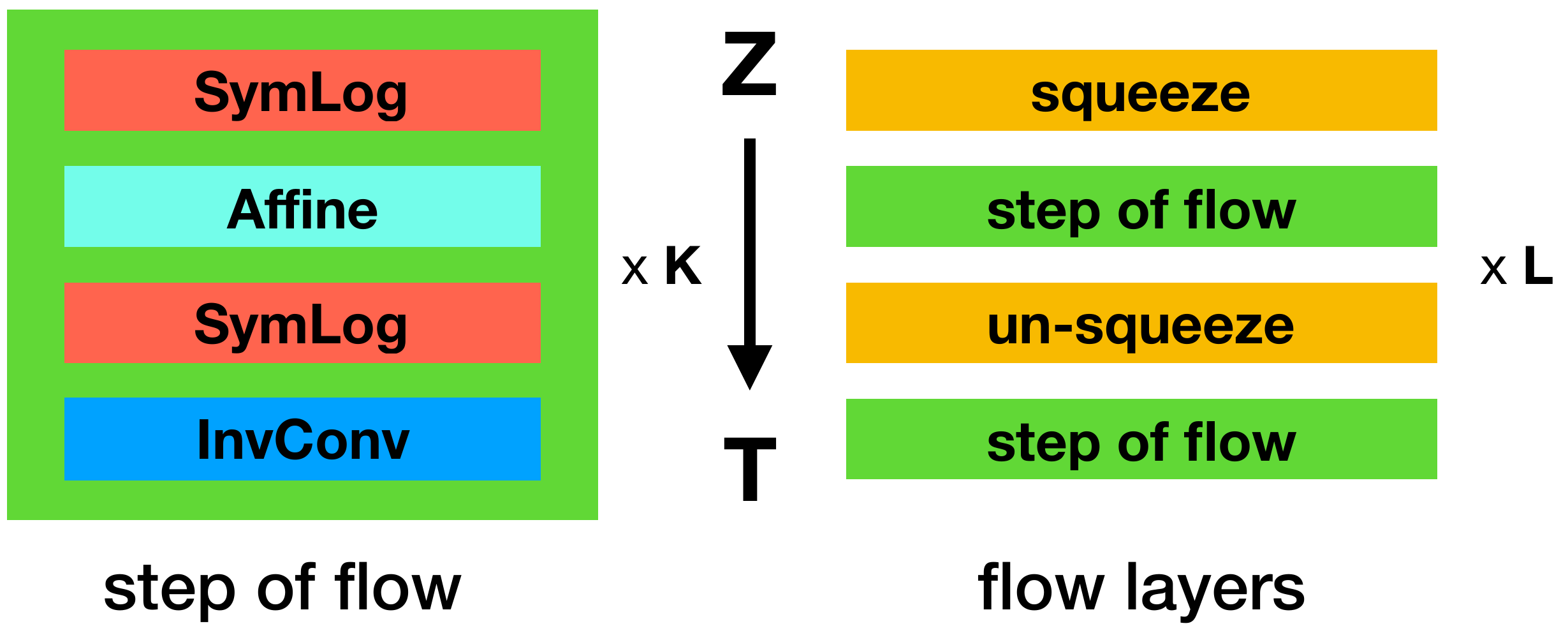}
    \\
    \vspace*{0.1cm}
    \textbf{(a)}  \TZK probability flow architecture
    \end{minipage}
    \begin{minipage}[b]{0.49\textwidth}
    \centering
    \includegraphics[width=1.0\textwidth]{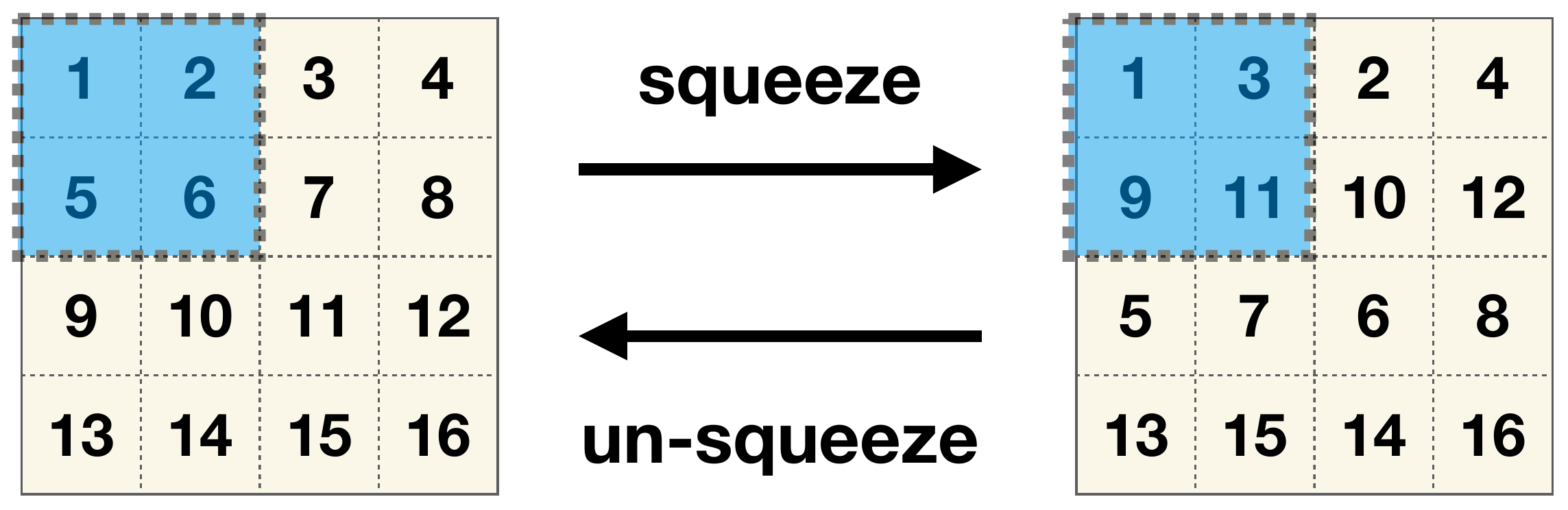}
    \\
    \vspace*{0.1cm}
    \textbf{(b)}  Tiling convolution
    \end{minipage}
    \caption{\TZK architecture. \textbf{(a)} We used channel-wise invertible convolution \cite{Kingma2018}, element-wise affine transformation \cite{Dinh2014}, and $SymLog\left(x\right) = \log\left(|x|+1\right) \cdot sign\left(x\right)$ as invertible activation. In order to propagate information between all dimensions we used squeeze/un-squeeze similar to \cite{Dinh2016a}.
    \textbf{(b)} Inspired by the versatility of CNN (\cite{Denker,Kingma2018}), we tiled the invertible convolution and alternated between squeeze and un-squeeze operations, allowing for reduced memory requirements.}
    \label{fig:tzk-architecture}
\end{figure}

\begin{figure}[t]
    \centering
    \begin{minipage}[b]{0.49\textwidth}
    \centering
    \scalebox{0.65}{
    \begin{tabular}{l|ll}
    \hline
                      & \textbf{MNIST} & \textbf{Omniglot} \\ \hline
    TzK (prior)       & 5.14      &  5.14    \\
    TzK (dataset conditional) $\ddagger$ & 3.56   &  3.75        \\
    TzK (digit conditional) $\dagger$ & 2.34   &  -        \\
    NICE              & 4.39  &      -    \\
    MADE MoG (conditional) \cite{Papamakarios2017} $\dagger$         & 1.39  &      -    \\
    MAF MoG (5) (conditional) \cite{Papamakarios2017} $\dagger$ & 1.51  &      -    \\
    Real NVP (5) \cite{Dinh2016a} $\dagger$ & 1.94  &      -    \\
    \end{tabular}}
    \\
    \vspace*{0.3cm}
    \textbf{(a)} Bits per pixel for conditional probability density estimation (lower is better) \cite{Papamakarios2017}.
    Label supervision is marked with $\dagger$, and dataset supervision with $\ddagger$.
    \end{minipage}
    \begin{minipage}[b]{0.49\textwidth}
    \centering
    \includegraphics[width=0.1\textwidth]{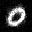}
    \includegraphics[width=0.1\textwidth]{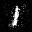}
    \includegraphics[width=0.1\textwidth]{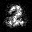}
    \includegraphics[width=0.1\textwidth]{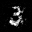}
    \includegraphics[width=0.1\textwidth]{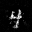}\\
    \includegraphics[width=0.1\textwidth]{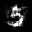}
    \includegraphics[width=0.1\textwidth]{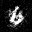}
    \includegraphics[width=0.1\textwidth]{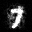}
    \includegraphics[width=0.1\textwidth]{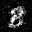}
    \includegraphics[width=0.1\textwidth]{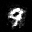}
    \\
    \vspace*{0.3cm}
    \textbf{(b)}  Samples from digit conditional priors.
    \\
    \vspace*{1.05cm}
    \end{minipage}
    \caption{Experimental results with unsupervised $\Ci \in \mathbb{R}^{2}$. \textbf{(a)} Negative log-likelihood bits per dimension, with (prior) refers to a non-conditional prior, (dataset conditional) refers to priors conditioned on dataset (MNIST or Omniglot), and (digit conditional) refers to priors conditioned on MNIST digit label (0-9). All other results are taken from \cite{Papamakarios2017}. \textbf{(b)} The digit conditional priors allows us to directly sample from manifold of interest in $\Tdomain$. \TZK formulation allows us to simultaneously train multiple conditional priors without hard-coding the number of priors (\eg, classes) in the architecture. See additional results in Fig.\ \ref{fig:tzk-additional-samples}.}
    \label{fig:tzk-experiments}
\end{figure}

We demonstrate \TZK by training jointly on MNIST and Omniglot datasets (see Fig.\ \ref{fig:tzk-experiments} and Fig.\ \ref{fig:tzk-additional-samples}).  \textbf{Architecture}: Priors are Gaussian, and conditional priors have additional 4 layers of flow from a conditional Gaussian  with NN regressors for mean and variance (4 linear layers with 70 units and PReLU). We used $L=10$ layers of flow step ($K=1$) with tiling convolution of size $16 \times 16$ as described in Fig. \ref{fig:tzk-architecture}. We used ADAM with $lr=1e-4$ for optimization. \textbf{Experiments:} We assign knowledge types per dataset (dataset conditional), and per MNIST label (digit conditional). We set $\Ci \in \mathbb{R}^2;\forall i$ to be unsupervised. \textbf{Preprocessing}: We dequantized pixel values for both datasets \cite{Peczenik1954,Li2014}, as described in \cite{Dinh2016a}, and padded MNIST to be $32 \times 32$. \textbf{Contributions}: We train a single model that allows us to turn a weak prior generative model into a powerful conditional generative model (Fig. \ref{sec:experiments}a). Compared with \cite{Dinh2016a,Papamakarios2017,Germain2015}, we did not have to encode a-priori the conditionals (number of classes) into the architecture, allowing \TZK to jointly train multiple conditional generative models, or incorporate new conditional generative models to an existing \TZK model in an on-line fashion.

\bibliographystyle{plain}
\bibliography{paper}

\newpage
\appendix
\section{Appendix} \label{sec:appendix}

\begin{figure}[h]
    \centering
    \begin{minipage}[b]{1.0\textwidth}
    \centering
    \includegraphics[width=1.0\textwidth]{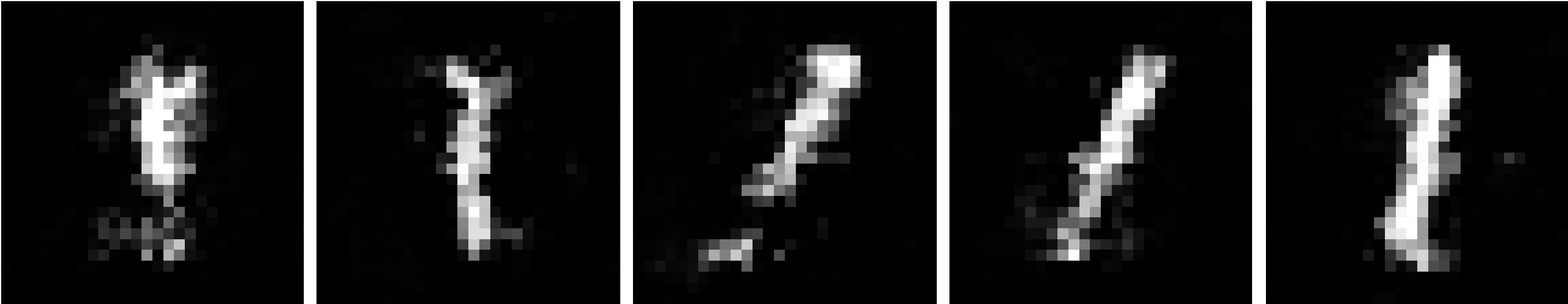}\\
    \includegraphics[width=1.0\textwidth]{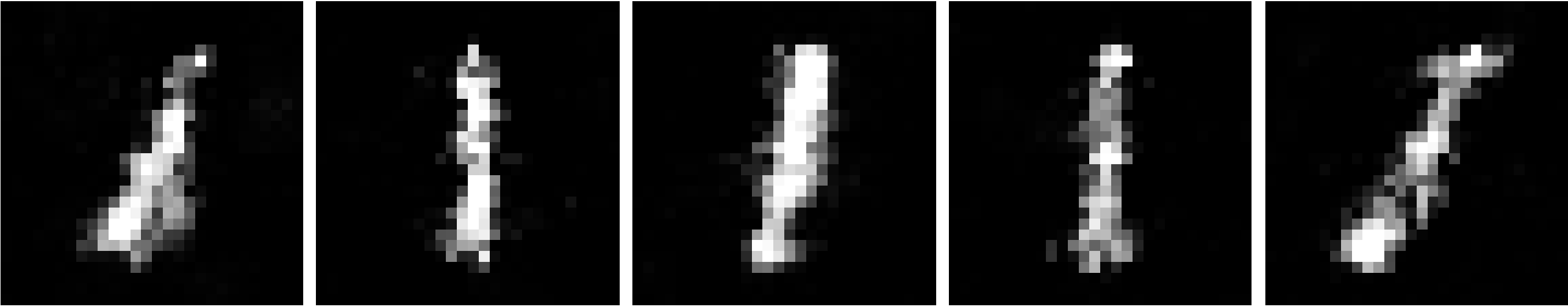}\\
    \textbf{(a)}  Samples from ``1'' conditional prior
    \end{minipage}

    \begin{minipage}[b]{1.0\textwidth}
    \centering
    \includegraphics[width=1.0\textwidth]{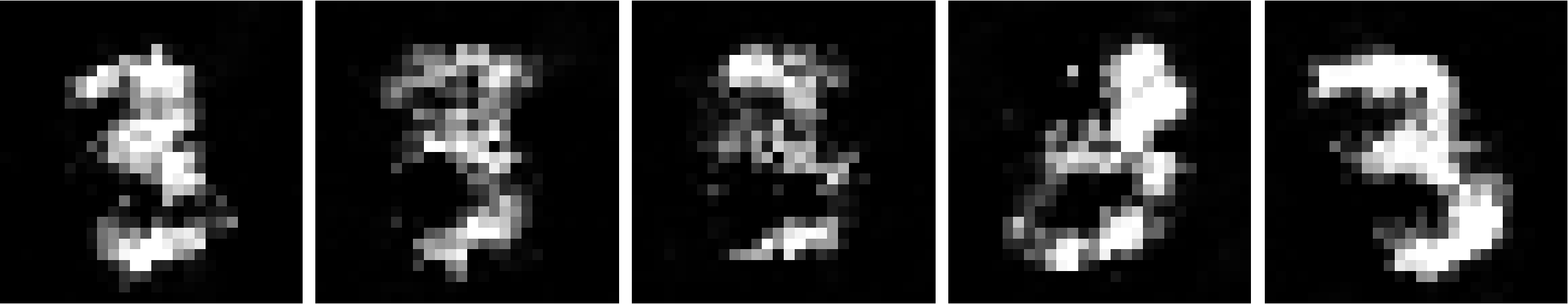}\\
    \includegraphics[width=1.0\textwidth]{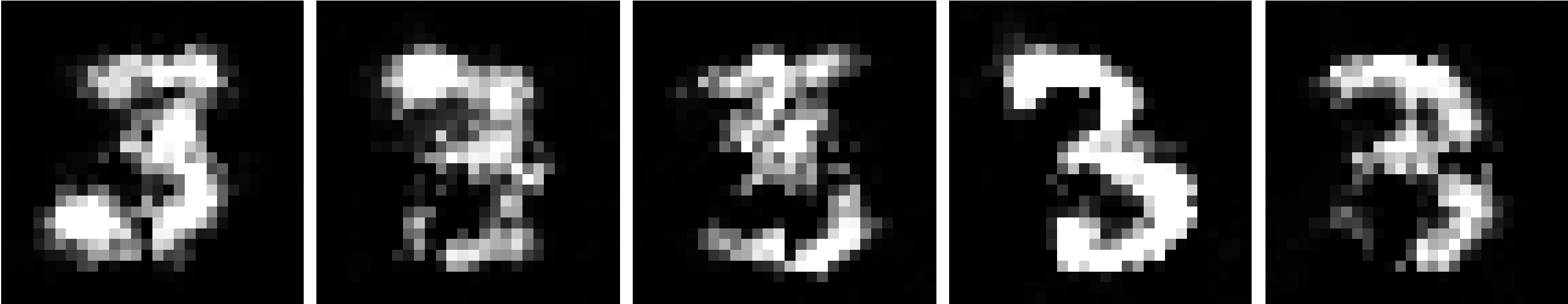}\\
    \textbf{(b)}  Samples from ``3'' conditional prior
    \end{minipage}

    \begin{minipage}[b]{1.0\textwidth}
    \centering
    \includegraphics[width=1.0\textwidth]{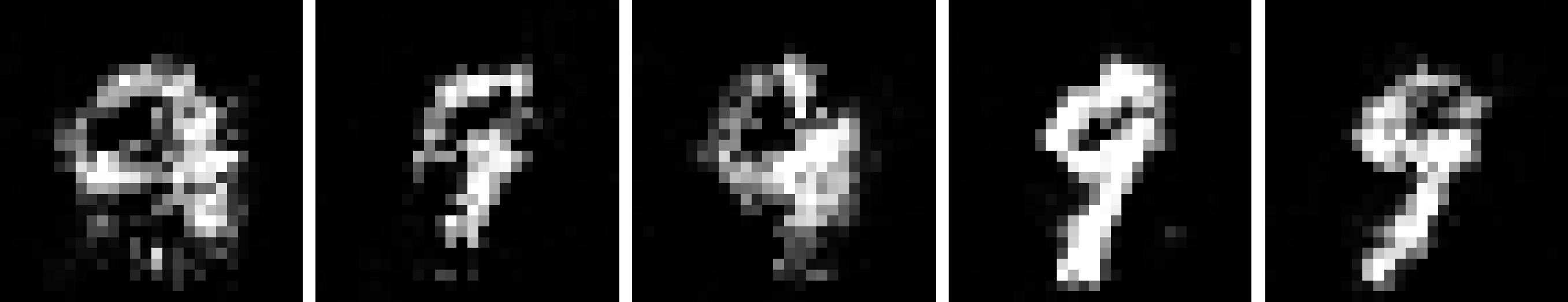}\\
    \includegraphics[width=1.0\textwidth]{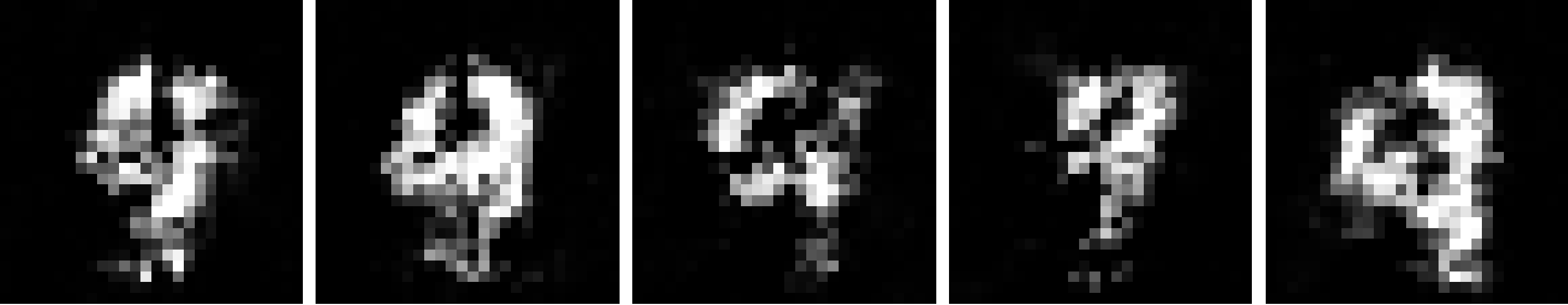}\\
    \textbf{(c)}  Samples from ``9'' conditional prior
    \end{minipage}

    \caption{Additional samples from MNIST label conditional priors with $\Ci \in \mathbb{R}^{2}$.}
    \label{fig:tzk-additional-samples}
\end{figure}

\end{document}